\documentclass{article}

\usepackage{arxiv}

\usepackage[utf8]{inputenc} 
\usepackage[T1]{fontenc}    
\usepackage{hyperref}       
\usepackage{url}            
\usepackage{booktabs}       
\usepackage{amsfonts}       
\usepackage{nicefrac}       
\usepackage{microtype}      
\usepackage{lipsum}
\usepackage{graphicx}
\usepackage{subfigure}
\usepackage{enumitem}
\usepackage{booktabs}
\usepackage{caption}
\usepackage{pgfplots}
\usepackage{tikz}

\title{Noisy Text Data: Achilles' Heel of BERT}

\author{
  Ankit Kumar\\
   \And
 Piyush Makhija \\
  \AND
  Anuj Gupta
 \AND
 Vahan Inc\\
 firstname@vahan.co\\
}

\begin{document}
\maketitle

\begin{abstract}
Owing to the phenomenal success of BERT on various NLP tasks and benchmark datasets, industry practitioners are actively experimenting with fine-tuning BERT to build NLP applications for solving industry use cases. For most datasets that are used by practitioners to build industrial NLP applications, it is hard to guarantee absence of any noise in the data. While BERT has performed exceedingly well for transferring the learnings from one use case to another, it remains unclear how BERT performs when fine-tuned on noisy text. In this work, we explore the sensitivity of BERT to noise in the data. We work with most commonly occurring noise (spelling mistakes, typos) and show that this results in significant degradation in the performance of BERT. We present experimental results to show that BERT's performance on fundamental NLP tasks like sentiment analysis and textual similarity drops significantly in the presence of (simulated) noise on benchmark datasets viz. IMDB Movie Review, STS-B, SST-2. Further, we identify shortcomings in the existing BERT pipeline that are responsible for this drop in performance. Our findings suggest that practitioners need to be vary of presence of noise in their datasets while fine-tuning BERT to solve industry use cases. 
\end{abstract}


\section{Introduction}
Pre-trained contextualized language models such as BERT (Bidirectional Encoder Representations from Transformers)~\cite{devlin2018bert}, which is the focus of this work, has led to improvement in performance on many Natural Language Processing (NLP) tasks. Without listing down all the tasks, BERT has improved the state-of-the-art for a number of tasks including tasks such as summarization, Name-Entity Recognition, Question Answering and Machine Translation~\cite{devlin2018bert}. 

Buoyed by this success, machine learning teams in industry are actively experimenting with fine-tuning BERT on their data to solve various industry use cases. These include use cases such as chatbots, sentiment analysis systems, automatically routing and prioritizing customer support tickets, NER systems, machine translation systems to name a few. Many of these use cases require practitioners to build training and test datasets by collecting text data from data sources \& applications such as chats, emails, discussions from user forums, social media conversations, output of machine translation systems, automatically transcribing text from speech data, automatically recognized text from printed or handwritten material, etc. Owing to these sources \& applications, the text data is known to be noisy. In some sources such discussions from user forums \& social media conversations, the noise in the data can be significantly high. \textit{It is not very clear how the pre-trained BERT performs when fine-tuned with noisy text data and if the performance degrades then why so}. These two questions are the focus of this paper.


Though the datasets used in industry are varied, many of them have a common characteristic - noisy text. This includes spelling mistakes, typographic errors, colloquialisms, abbreviations, slang, internet jargon, emojis, embedded metadata (such as hashtags, URLs, mentions), non standard syntactic constructions and spelling variations, grammatically incorrect text,  mixture of two or more languages (a.k.a code mix) to name a few. This makes cleaning and preprocessing text data a key component of NLP pipeline for industrial applications. 
However, despite extensive cleaning and preprocessing, some degree of noise often remains.  Owing to this residual noise, a common issue that NLP models have to deal with is out of vocabulary (OOV) words. These are words that are found in test and production data but are not part of training data. In this work  we find that BERT fails to properly handle OOV words (due to noise). We show that this negatively impacts the performance of BERT on fundamental tasks in NLP when fine-tuned over noisy text data. 

This work is motivated from the business use case where we are building a dialogue system over WhatsApp to screen candidates for blue collar jobs. Our candidate user base often comes from underprivileged backgrounds, many of them are not even college graduates. This coupled with fat finger problem\footnote{https://en.wikipedia.org/wiki/Fat-finger\_error} over a mobile keypad leads to a lot of typos and spelling mistakes in the responses sent to our dialogue system. Hence, for the purpose of this work, we focus on spelling mistakes as the noise in the data. While this work is motivated from our business use case, our findings are applicable to other use cases that deal with noisy text data.

\section{Previous Work}

We now present some of the relevant work viz the following related areas viz. (1) robustness of BERT, (2) degradation in performance of NLP models due to noise in text data. 

\vspace{10pt}
\textbf{Robustness of BERT}: There has been some work on testing the robustness of BERT in different scenarios. \cite{jin2019bert} introduce TEXTFOOLER, a system to generate adversarial text and apply it to text classification and textual entailment to successfully attack the pre-trained BERT among other models.\cite{aspillaga2020stress} evaluate robustness of three models - RoBERTa, XLNet, and BERT in Natural Language Inference (NLI) and Question Answering (QA) tasks. They show that while RoBERTa, XLNet and BERT are more robust than Recurrent Neural Network (RNN) models to stress tests on tasks such as NLI and QA, these models are still very fragile and show many unexpected behaviors. \cite{pal2020transfer} present novel attack techniques that utilize the unintended features learnt in the teacher (public) model to generate adversarial examples for student (downstream) models. They show that using length-based and sentence-based misclassification attacks for the Fake News Detection task trained using a context-aware BERT model, one gets misclassification accuracy of 78\% and 39\% respectively for the adversarial examples. \cite{Sun2020AdvBERTBI} show the BERT under-performs on sentiment analysis and question answering in presence of typos and spelling mistakes. While our work has an overlap with their work, our work is independent (and parallel in terms of timeline) to their work. 

We not only experimented with more datasets, we also pin down the exact reason for degradation in BERT's performance. We demonstrate our findings viz-a-viz two most fundamental NLP tasks - sentence classification (sentiment analysis) and textual similarity. For these, we chose the most popular benchmark datasets - for sentiment analysis we work with SST-2 and IMDB datasets and for textual similarity we use STS-B dataset. Further, \cite{Sun2020AdvBERTBI} show that mistakes/typos in the most informative words cause maximum damage. In contrast, our work shows stronger results - mistakes/typos in words chosen at random is good enough to cause substantial drop in BERT's performance. We discuss the reason for performance degradation. Last but not the least, we tried various tokenizers during the fine-tuning phase to see if there is a simple fix for the problem.  


\vspace{10pt}
\noindent \textbf{Degradation in performance of NLP models due to Noise}: There has been a lot of work around understanding the effect of noise on the performance of NLP models.\cite{taghva2000evaluating} evaluate the effect of OCR errors on text categorization. \cite{wu2016google} introduced ISSAC, a system to clean dirty text from online sources. \cite{agarwal2007much} studied the effect of different kinds of noise on automatic text classification. \cite{Subramaniam2009} presented a survey of types of text noise and techniques to handle noisy text. Newer communication mediums such as SMS, chats, twitter, messaging apps encourage brevity and informalism, leading to non-canonical text. This presents significant challenges to the known NLP techniques. \cite{belinkov2017synthetic} show that character based neural machine translation (NMT) models are also prone to synthetic and natural noise even though these model do better job to handle out-of-vocabulary issues and learn better morphological representation. 
\cite{ribeiro2018semantically} develop a technique, called semantically equivalent adversarial rules (SEARs) to debug NLP models. SEAR generate adversial examples to penetrate NLP models. Author experimented this techniques for three domains: machine comprehension, visual question answering, and sentiment analysis.

There exists a vast literature that tries to understand the sensitivity of NLP models to noise and develop techniques to tackle these challenges. It is beyond the scope of this paper to give a comprehensive list of papers on this topic. One can look at the work published in conferences such as 'Workshop on Noisy User-generated Text, ACL', 'Workshop on Analytics for Noisy Unstructured Text Data, IJCAI-2007' that have dedicated tracks on these issues.

\section{Experiments}

We evaluate the state-of-the-art model BERT\footnote{BERT\textsubscript{Base} uncased model} on two fundamental NLP tasks: sentiment analysis and textual similarity. For sentiment analysis we use popular datasets of IMDB movie reviews \cite{maas2011learning} and Stanford Sentiment Treebank (SST-2)~\cite{socher2013recursive}; for textual similarity we use Semantic Textual Similarity (STS-B) \cite{cer2017semeval}. Both STS-B and SST-2 datasets are a part of GLUE benchmark \cite{wang2018glue} tasks. On these benchmark datasets we report the system's performance both - with and without noise. 

\subsection{Noise}

As mentioned in Section 1 of the paper, we focus on the noise introduced by spelling mistakes and typos. All the benchmark datasets we work with consists of examples X \(\rightarrow\) Y where X are the text inputs and Y are the corresponding labels. We call the original dataset as D\textsubscript{0}. From D\textsubscript{0} we create new datasets D\textsubscript{2.5}, D\textsubscript{5}, D\textsubscript{7.5}, D\textsubscript{10}, D\textsubscript{12.5}, D\textsubscript{15}, D\textsubscript{17.5}, D\textsubscript{20} and D\textsubscript{22.5}. Here, D\textsubscript{k} is a variant of D\textsubscript{0} with k\% noise in each datapoint in D\textsubscript{0}. 
 
To create D\textsubscript{k}, we take i\textsuperscript{th} data point x\textsubscript{i} \(\in\) D\textsubscript{k}, and introduce noise in it. We represent the modified datapoint by x\textsubscript{i,k}\textsuperscript{noise} . Then, D\textsubscript{k} is simply the collection (x\textsubscript{i,k} \textsuperscript{noise}, y\textsubscript{i}), \(\forall\)i. To create x\textsubscript{i,k}\textsuperscript{noise} from x\textsubscript{i}, we randomly choose k\% characters from the text of x\textsubscript{i} and replace them with nearby characters in a qwerty keyboard. For example, if character \emph{d} is chosen, then it is replaced by a character randomly chosen from \emph{e}, \emph{s}, \emph{x}, \emph{c}, \emph{f}, or \emph{r}. This is because in a qwerty keyboard, these keys surround the key \emph{d}. We  inject noise in the complete dataset. Later we split D\textsubscript{i} into \textit{train} and \textit{test} chunks. 

We believe a systematic study should be done to understand how the performance of SOTA models is impacted when fine-tuned on noisy text data. To motivate the community for this, we suggest a simple framework for the study. The framework uses four variables -  SOTA model, task, dataset, and the degree of noise in the data. For such a study it is imperative to have a scalable way to create variants of a dataset that differ in the degree of noise in them. The method for creating noisy datasets as described in the previous paragraph does exactly this. Creating datasets at scale with varying degree of natural noise is very human intensive task. Despite our method introducing synthetic noise, owing to mobile penetration across the globe, and fat finger problem, our noise model is very realistic. Also, unlike \cite{Sun2020AdvBERTBI}, we introduce noise randomly rather than targeting the most informative words. This helps us model the average case setting rather than the worst case. For these reasons we stick to synthetic noise introduced randomly.





\subsection{Sentiment Analysis}
For sentiment analysis we use IMDB movie reviews \cite{maas2011learning} and Stanford Sentiment Treebank (SST-2 ) \cite{socher2013recursive} datasets in binary prediction settings. IMDB datasets consist of 25000 training and 25000 test sentences. We represent the original IMDB dataset (one with no noise) as IMDB\textsubscript{0}. Using the process of introducing noise (as described in section 3.1), we create 9 variants of IMDB\textsubscript{0} namely IMDB\textsubscript{2.5}, \ldots, IMDB\textsubscript{22.5} with varying degrees of noise. 

SST-2 dataset consists of 67349 training and 872 test sentences. Here too we we add noise as described in Section 3.1 to create 9 variants of SST-2\textsubscript{0} -  SST-2\textsubscript{2.5}, \ldots,  SST-2\textsubscript{22.5}.
To measure the performance of the model for sentiment analysis task we use F1 score.

\subsection{Textual Similarity}

For textual similarity task, we use Semantic Textual Similarity (STS-B) \cite{cer2017semeval} dataset. The dataset consists of 5749 training and 1500 test data points. Each data point consists of 2 sentences and a score between 0-5 representing the similarity between the two sentences. We represent the original data set by STS-B\textsubscript{0} and create 9 noisy variants like we mentioned in section 3.1
Here, we use Pearson-Spearman correlation to measure model's performance.

\subsection{Results}
 
\begin{figure}
\centering
\begin{tikzpicture}
\begin{axis}[
    title={Results of experiments on both clean and noisy data},
    xlabel={\% error},
    ylabel={Accuracy},
    xmin=0, xmax=22.5,
    ymin=0, ymax=1,
    xtick={0,2.5,5,7.5,10.0,12.5,15,17.5,20,22.5},
    ytick={0.1,0.2,0.3,0.4,0.5,0.6,0.7,0.8,0.9,1.0},
    legend pos=south west,
    ymajorgrids=true,
    grid style=dashed]

\addplot[
    color=blue,
    mark=square,
    ]
    coordinates {
    (0,0.93)(2.5,0.85)(5,0.70)(7.5,0.67)(10,0.62)(12.5,0.53)(15,0.51)(17.5,0.46)(20,0.44)(22.5,0.41)
    };

\addplot[
    color=red,
    mark=square,
    ]
    coordinates {
    (0,0.89)(2.5,0.86)(5,0.80)(7.5,0.76)(10,0.70)(12.5,0.67)(15,0.60)(17.5,0.59)(20,0.54)(22.5,0.49)
    };

\addplot[
    color=green,
    mark=square,
    ]
    coordinates {
    (0,0.89)(2.5,0.84)(5,0.75)(7.5,0.65)(10,0.65)(12.5,0.49)(15,0.40)(17.5,0.39)(20,0.29)(22.5,0.31)
    };
    \legend{IMDB,SST-2,STS-B}

\end{axis}

\end{tikzpicture}
\captionsetup{justification=centering}
\caption{Accuracy vs Error}
\label{figure:resultsOnNoisyData}
\end{figure}
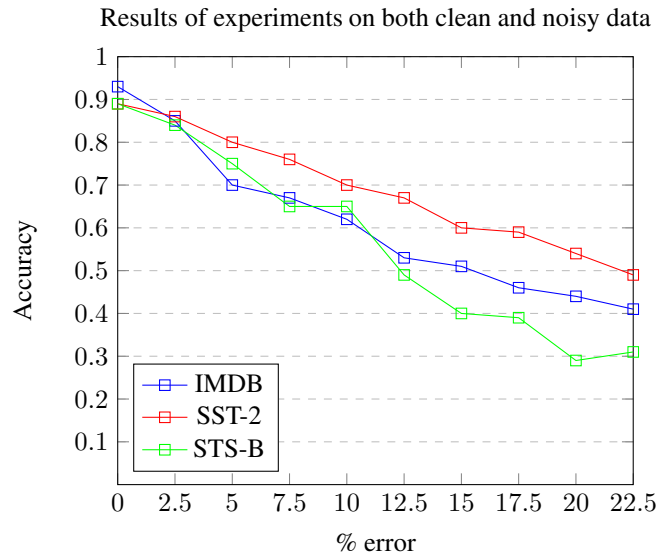

Table~\ref{table:resultsOnNoisyData} and Figure~\ref{figure:resultsOnNoisyData}  lists the performance of BERT on various variants (noiseless and noisy) of IMDB and STS-2 for sentiment analysis and SST-B for sentence similarity. From the numbers it is very clear that noise adversely affects the performance of BERT. Further, as we gradually increase the noise, the performance keeps going down. For sentiment analysis, by the time 15-17\% of characters are replaced, the performance drops to almost the chance-level accuracy (i.e., around 50\%). This decline is much more rapid for sentence similarity.

\section{Analysis}

To understand the reason behind the drop in BERT's performance in presence of noise, we need to understand how BERT processes input text data. A key component of BERT's pipeline is tokenization of input text. It first performs whitespace tokenization followed by WordPiece tokenization \cite{wu2016google}. While whitespace tokenizer breaks the input text into tokens around the whitespace boundary, the wordPiece tokenizer uses longest prefix match to further break the tokens\footnote{https://github.com/google-research/bert/blob/master/tokenization.py}. The resultant tokens are then fed as input to the BERT model.

\begin{table}[]
\begin{center}

\begin{tabular}{|c|c|c|c|}
\hline
\textbf{}         & \multicolumn{2}{l|}{\textbf{Sentiment Analysis}} & \textbf{Textual Similarity} \\
\hline
\textbf{\% error} & \textbf{IMDB}          & \textbf{SST-2}          & \textbf{STS-B}              \\
\hline
0.0                 & 0.93                   & 0.89                    & 0.89                        \\
2.5               & 0.85                   & 0.86                    & 0.84                        \\
5.0                 & 0.79                   & 0.80                    & 0.75                        \\
7.5               & 0.67                   & 0.76                    & 0.65                        \\
10.0                & 0.62                   & 0.70                    & 0.65                        \\
12.5              & 0.53                   & 0.67                    & 0.49                        \\
15.0                & 0.51                   & 0.60                    & 0.40                        \\
17.5              & 0.46                   & 0.59                    & 0.39                        \\
20.0                & 0.44                   & 0.54                    & 0.29                        \\
22.5              & 0.41                   & 0.49                    & 0.31  
\\
\hline
\end{tabular}
\end{center}
\vspace{10pt}
\caption{Results of experiments on both clean and noisy data. }
\label{table:resultsOnNoisyData}
\end{table}

When it comes to tokenizing the noisy text data, we see a very interesting behaviour from BERT's pipeline. First whitespace tokenization is applied. Now, when the WordPiece tokenizer encounters these words, owing to the spelling mistakes, these words are not directly found in BERT's dictionary. So, WordPiece tokenizer tries to tokenize these (noisy) words into subwords. However, \textit{it ends up breaking the words into subwords whose meaning can be  very different from the meaning of the original word}. \textit{This can change the meaning of the sentence completely, therefore leading to substantial dip in the performance}. 

To understand this better let us look at two examples - one each from the IMDB and STS-B datasets respectively, as shown in Example 1 and Example 2. In each Example, (a) is the sentence as it appears in IMDB\textsubscript{0} (i.e. original dataset) while (b) is the corresponding sentence after adding 5\% noise (IMDB\textsubscript{5}). For legibility the misspelled characters are highlighted with italics. The sentences are followed by their corresponding output after applying whitespace and WordPiece tokenizer on them. In the output, \#\# represents subwords. 
 \newline

\noindent \textbf{Example 1 (from IMDB)}:
    
\noindent  (a) that loves its characters and communicates something rather beautiful about human nature. (0\% error)

 \noindent (b) that loves \textit{8}ts characters a\textit{b}d communicates something rathe\textit{e} beautiful about human natu\textit{e}e. (5\% error)
 
\noindent \textbf{Corresponding output of tokenization}:
       
 \noindent (a) `that', `loves', `its', `characters', `and', `communicate', 
   `\#\#s', `something', `rather', `beautiful', `about', `human',     
   `nature'
  
 \noindent (b) `that', `loves', `8', `\#\#ts', `characters', `abd', `communicate', 
        `\#\#s', `something', `rat', `\#\#hee', `beautiful', `about', `human', 
        `nat', `\#\#ue', `\#\#e'   \newline
 
\noindent \textbf{Example 2 (from STS-2)}: 
 
\noindent (a) poor ben bratt could n't find stardom if mapquest emailed him point-to-point driving directions. (0\% error)
 
\noindent (b) poor ben bratt could n't find stardom if mapquest emailed him point-to-point drivi\textit{b}g dir\textit{s}ctio\textit{je}. (5\% error)
   
 \noindent \textbf{Output of tokenization}:
 
\noindent (a) `poor', `ben', `brat', `\#\#t', `could', `n', `'', `t', `find',  `star', `\#\#dom', `if', `map', `\#\#quest', `email', '\#\#ed', `him', `point', `-', `to', `-', `point', `driving', `directions', `.' 
 
\noindent (b) `poor', `ben', `brat', `\#\#t', `could', `n', `'', `t', `find', `star', `\#\#dom', `if', `map', `\#\#quest', `email', `\#\#ed', `him',  `point', `-', `to', `-', `point', `dr', `\#\#iv', `\#\#ib', `\#\#g',  `dir', `\#\#sc', `\#\#ti', `\#\#oge', `.'\\

 In  Example 1(a), BERT's tokenization splits \emph{communicates} into \emph{communicate} and \emph{\#\#s} based on longest prefix matching because there is no exact match for \emph{communicates} in pre-trained BERT's vocabulary. This results in two tokens \emph{communicate}  and \emph{s},  both of which are present in BERT's vocabulary. We have contextual embeddings for both \emph{communicate} and \emph{\#\#s}. By using these two embeddings, one can get an approximate embedding for \emph{communicates}. 
 
However, this approach goes for a complete toss when the word gets misspelled. 
In example 1(b), the word \emph{natuee}('nature' is misspelled) is split into tokens \emph{nat, \#\#ue, \#\#e} (based on the longest prefix match). By combining the embeddings for these three tokens, one cannot approximate the embedding of 'nature'. This is because the word \textit{nat} has a very different meaning (it means 'a person who advocates political independence for a particular country'). This misrepresentation in turn impacts the performance of downstream sub-components of BERT bringing down the overall performance of BERT model. This is why as we introduce more errors, the quality of output of the tokenizer degrades further, resulting in the overall drop in performance.  
 
\begin{table}
\begin{center}
\begin{tabular}{|c|c|c|c|}
\hline
\textbf{\% error} & \textbf{WordPiece} & \textbf{WhiteSpace} & \textbf{n-gram} \\
              &              &             & (n=6)  \\
\hline
0              & 0.89               & 0.69                & 0.73    \\
5              & 0.75               & 0.59                & 0.60                 \\
10             & 0.65               & 0.41                & 0.46                 \\
15             & 0.40               & 0.33                & 0.36                 \\
20             & 0.39               & 0.22                & 0.25                
\\
\hline

\end{tabular}
\end{center}
\vspace{10pt}
\centering
\caption{Comparative results on STS-B dataset with different tokenizers}
\label{table:resultsWithOtherTokenizer}
\end{table}
%

We further experimented with different tokenizers other than WordPiece tokenizer. For this we used Character N-gram tokenizer \cite{mcnamee2004character} and stanfordNLP whitespace tokenizer \cite{manning2014stanford}. For Character N-gram tokenizer, we work with N=6\footnote{longer N-grams such as 6-grams are recommended for capturing semantic information \cite{fasttext}}. The results of these experiments on STS-B dataset are given in Table~\ref{table:resultsWithOtherTokenizer}. It is clear that replacing WordPiece by whitespace or N-gram further degrades the performance. The reasons are as follows: 

(1) \textbf{Replace WordPiece by whitespace}: In this case every misspelled word (say, natuee) is directly fed to BERT model. Since, these words are not present in BERT's vocabulary, they are treated as UNK\footnote{Unknown tokens} token. In presence of even 5-10\% noise, there is a significant drop in accuracy. 

(2) \textbf{Replace WordPiece by Character N-gram tokenizer}: Here, every misspelled word is broken into character n-grams of length atmost 6. It is high unlikely to find these subwords in BERT's vocabulary. Hence, they get treated as UNK. 

Please note that in our experimental setup, we are not training BERT from scratch. Instead, we simply replaced the existing WordPiece tokenizer with other tokenizers while feeding tokens to BERT's embedding layer during the fine-tuning and testing phases.

\section{Conclusion and Future Work}
We studied the effect of synthetic noise (spelling mistakes) in text data on the performance of BERT. We demonstrated that as the noise increases, BERT's performance drops drastically. We further show that the reason for the performance drop is how BERT's tokenizer (WordPiece) handles the misspelled words. 

Our results suggest that one needs to conduct a large number of experiments to see if the findings hold across other datasets and popular NLP tasks such as information extraction, text summarization, machine translation, question answering, etc. It will also be interesting to see how BERT performs in presence of other types of noise. One also needs to investigate how other models such as ELMo, RoBERTa, and XLNet which use character-based, byte-level BPE, and SentencePiece tokenizers respectively. It also remains to be seen if the results will hold if the noise was restricted to only frequent misspellings. 

To address the problem of drop in performance, there are 2 ways -  (i) preprocess the data to correct spelling mistakes in the dataset before fine-tuning BERT on it (ii) make changes in BERT's architecture to make it robust to noise. From a practitioner's perspective, the problem with (i) is that in most industrial settings this becomes a separate project in itself. We leave (ii) as future work.

\bibliographystyle{unsrt}  

\bibliography{bertugc}

\end{document}